\documentclass[sigconf]{acmart}
\usepackage[utf8]{inputenc}
\usepackage[T1]{fontenc} 
\usepackage{tikz}
\usetikzlibrary{fit,positioning,bayesnet,shapes,arrows}
\usepackage{url}
\usepackage{listings}
\usepackage{multirow}
\usepackage{changepage}
\usepackage{comment}
\newcommand{\bg}[1]{ \boldsymbol{#1} }
\usepackage{graphicx}

\setcopyright{rightsretained}
\copyrightyear{2017}
\acmYear{2017}
\setcopyright{acmcopyright}
\acmConference{KDD '17}{August 13-17, 2017}{Halifax, NS,Canada}
\acmPrice{15.00}
\begin{document}

\title{Luck is Hard to Beat: The Difficulty of Sports Prediction} 
\author{Raquel Aoki}
\affiliation{
\institution{Department of Computer Science}
\streetaddress{UFMG}
\city{Belo Horizonte}
\country{Brazil}
}
\email{raquel.aoki@dcc.ufmg.br}

\author{Renato Assunção}
\affiliation{
\institution{Department of Computer Science}
\streetaddress{UFMG}
\city{Belo Horizonte}
\country{Brazil}
}
\email{assuncao@dcc.ufmg.br}

\author{Pedro Vaz de Melo}
\affiliation{%
\institution{Department of Computer Science}
\streetaddress{UFMG}
\city{Belo Horizonte}
\country{Brazil}
}
\email{olmo@dcc.ufmg.br}

\begin{abstract}
Predicting the outcome of sports events is a hard task. We quantify this difficulty with a coefficient that measures the distance between the observed final results of sports leagues and idealized perfectly balanced competitions in terms of skill. This indicates the relative presence of luck and skill. We collected and analyzed all games from 198 sports leagues comprising 1503 seasons from 84 countries of 4 different sports: basketball, soccer, volleyball and handball. We measured the competitiveness by countries and sports. We also identify in each season which teams, if removed from its league, result in a completely random tournament. Surprisingly, not many of them are needed. As another contribution of this paper, we propose a probabilistic graphical model to learn about the teams' skills and to decompose the relative weights of luck and skill in each game. We break down the skill component into factors associated with the teams' characteristics. The model also allows to estimate as 0.36 the probability that an underdog team wins in the NBA league, with a home advantage adding 0.09 to this probability. As shown in the first part of the paper, luck is substantially present even in the most competitive championships, which partially explains why sophisticated and complex feature-based models hardly beat simple models in the task of forecasting sports' outcomes.
\end{abstract}




\maketitle

\section{Introduction}

Millions of fans watched mesmerized the Super Bowl 2017 outcome. The New England Patriots had a spectacular 
comeback defeating the Atlanta Falcons in a historical game. When the Falcons opened up a 28-3 lead at 
the start of the third quarter, no one would bet on the Patriots. And yet they won in an amazing way. 
After the game, endless discussion tables ponder over whether the Falcons blew it or the Patriots won it. 
Was it luck or skill that played the main role in that evening? Either luck or skill, to watch a solid victory 
melting and dissolving in the air brings the feeling that the world is unpredictable and uncontrollable. 

There have been empirical and theoretical studies showing that unpredictability cannot 
be avoided~\cite{Watts2016}. 
Indeed, Martin et al.~\cite{Watts2016} showed that realistic bounds on predicting outcomes in social systems 
imposes drastic 
limits on what the best performing models can deliver. And yet, accurate prediction is a holy grail 
avidly sought 
in financial markets~\cite{chen2009}, sports~\cite{chen2016predicting}, 
arts and entertainment award events~\cite{haughton2015oscar}, 
and politics~\cite{tumasjan2010}. 

Among all these domains, sports competitions appears as a specially attractive field of study for three main reasons.
The first one is that a sports league forms a relatively isolated system, with few external and unrestrained influences. 
Furthermore, it is replicated over time approximately in the same conditions and under the same rules. 
The second reason is the large 
amount of data available that makes possible to learn their statistical patterns. The third reason is their popularity, 
which fuels a multibillion business that includes television, advertisement, and a huge betting market. All these 
business players would benefit from better understanding of predictability in sports. 

Sports outcomes is a mix of skills and good and bad luck~\cite{Owen2013, fortmaxcy2003, Zimbalist2002}. This mix is responsible for large part of the sports appeal~\cite{Chan2009}. 
While an individual player may find fun equally with a pure skill game (such as chess) 
or with a pure luck game (such as a lottery), for the typical large sports audience the mix aspect is the 
great attraction. If a pure skill game, it would be completely predictable and potentially 
boring~\cite{fort2011}. 
There is also the suspition among researchers that a competition based on pure luck will 
lower audience interest and hence, less revenue~\cite{Owen2013}. 
If a pure luck game, the final champions will not be deemed worthy of admiration as their ranks are not resulting 
their merits and  not enough positive psychological emotions will be aroused~\cite{khanin2000emotions}.  
Thus, the mixture makes prediction feasible, but very hard. According to Blastland and Dilnot~\cite{numbersgame}, 
the teams favored by bettors win half the time in soccer, 60\% of the time in baseball,
and 70\% of the time in football and in basketball. Despite the large amount of money involved,
there are no algorithms capable of producing accurate predictions and there is some evidence they 
will never be found~\cite{Watts2016}.

Thus, the objective of this paper is to propose a general framework to measure the role of skill in sports leagues' outcomes. In order to do that, we propose a hypothesis test to verify if skill actually plays a significant role in the league. If it does, our framework identifies which teams are significantly more and less skilled than others. Finally, for leagues in which skill plays a significant role, we present a probabilistic graphical model to learn about the teams' skills.

As the first contribution of this paper, 
we introduce the \textit{skill} coefficient, denoted by $\phi$, that measures the 
departure between the observed final season score distribution and what one could expect from a competition 
based on pure luck plus contextual circumstances. The larger
the value of $\phi$, the more distant from a no-luck competition. From $\phi$, we devise two techniques 
to characterize the role of skill in sports leagues. First, we propose a significance test for $\phi$, 
which shows, for instance, that most leagues' outcomes in 
certain sports, such as handball, are no distinct from a random ordering of the teams. 
Second, we propose a technique based on $\phi$ that identifies which teams are significantly more (and less) skilled than the others in the league. 
By this technique, it is possible to show, for example, that in leagues where the competition is intense and 
the financial stakes are very high, such as the \emph{Primera Divisi\'{o}n}
Spanish soccer league, the removal of only 3 teams out of 20 renders the competition completely random. 
We also show through the $\phi$ coefficient the unequivocal presence of rare (or anomalous) phenomena in some leagues. 
In these cases, the final scores end up very close to each other, with very little difference between the teams.
We show that this similarity is so extreme that there is almost no possibility that it could have been 
generated by a completely random competition, let alone a competition where different skills are present.   

The value of $\phi$ increases with the weight of the skill component but this relationship is unknown and nonlinear. 
It does not answer a most relevant question: how much of the observed variation 
in the final scores can be assigned to skill or luck? A large value of $\phi$ 
may be due to the presence of few outliers with extreme skills or it can be due to a distribution without 
outliers but with excessive kurtosis. Hence, our second contribution is a probabilistic graphical model 
that aims to disentangle and measure the relative weights of skills and randomness in the final scores. 
We assume a random effects model with a generalized linear model allowing for features to explain 
the skill differences between the teams. We model the large number of pairwise matchups in a sports competition 
with a generalization of the Bradley-Terry paired comparison model that uses latent skill factors and 
random effects due to the luck factor. Based on this graphical model, we are able to estimate the 
luck component in the competition and so answer to the question: by how much skill determines the result?
An additional result is that the analysis of the coefficients associated with the features 
suggest a number of management strategies to build a highly competitive team. 

As our final contribution, we present a deep and comprehensive characterization of the presence of luck and skill in sports leagues using our proposed $\phi$ coefficient. We collected and analyzed all games from 198 sports leagues of 4 different sports: basketball, soccer, volleyball and handball. In total, 1503 seasons from 84 different countries were analyzed. From this analysis, we revealed which sport is more likely to have a league whose outcomes could be completely explained by chance. Also, we showed, for each sport, what is the expected percentage of teams that should be removed from a sports league to make such league similar with pure luck competition. In this direction, by using $\phi$ over the historical outcomes of sports leagues, we characterize how stable (or dynamic) are the skills of teams in that league. 

\section{Related Work}

There is a large literature aiming at predicting the final outcome of individual sports games. 
Vra\v{c}ar \emph{et al.} \cite{Vracar2016} proposed an ingenious model based on Markov process 
coupled with a multinomial logistic regression approach to predict each consecutive point in a basketball match. 
A large number of features included the current context in the progression of a game. Hence, they used the 
game time, the points difference at that moment, as well as the opposing teams’ characteristics. 
Another model for the same situation of the sequence of successive points was proposed by \cite{peel2015} 
where the focus was to verify the influence of the leading gap size (is there a restoration force) 
and the identity of the last team to score (anti-persistence). 
\cite{Gabel2012} and \cite{Merritt2014} found that extremely simple stochastic models 
fitted the empirical data very well. The common pattern is that events occur randomly, 
according to a homogeneous Poisson process
with a sport-specific rate. Given an event, its points are assigned to one of the teams 
by flipping a game-specific biased coin (a Bernoulli process). 
Vaz de melo \cite{vaz2012forecasting} proposed features to represent the network 
effects between the players on the teams performance. 
In \cite{ben2006parity}, the authors fit a idealized and simple theoretical model to empirical 
data in order to estimate a so-called upset probability $q$, the chance that a worse team wins.
Although simple, the model fits quite well to the data and they conclude that $q \approx 0.45$
in the case of soccer and baseball, while $q \approx 0.35$ for basketball and football. 
These numbers give a rough idea of the level of randomness present in the sports leagues. 
The within game prediction of each successive point is a hard task as the most successful 
models typically calculate the most extreme probability around 0.65 
for a given team to score next, conditional on the model features \cite{peel2015}.

Probabilistic graphical models have been used also. 
Chen and Joachims presented in \cite{chen2016predicting} a probabilistic framework for predicting 
the outcome of pairwise matchups considering contextual characteristics of the game, such as the 
weather or the type of tennis field.

Predicting the final ranking in a league is a different task but equally hard to perform well.
In this case, one is interested in estimating the chance that the strongest team is the final 
winner or the probability that a weak team ended up being the champion. 
Besides the within-game factors, there is also effects due to the competition 
design~\cite{Chan2009, ben2013randomness, BenNaimKorea}. 
Chetrite studied in \cite{chetrite2015number}  the number of potential winners in 
competitions  considering how their abilities is distributed. 
Spiegelhalter studied how measure the luck and ability in a championship in \cite{futebol1} by 
comparing the sample variance and the theoretical variance expected when all teams have the same ability. 
A ranking to NBA teams was made by Pelechrinis in \cite{pelechrinis2016sportsnetrank}. It was used the PageRank Algorithm to create a network among the teams. This work has an accuracy of about 67\%. 

The keen interest from industry and academia in sports have spur many recent papers aiming at guiding 
the tactical moves during the game. \cite{wang2015} discovered the best tactical patterns of soccer teams
by mining the historical match logs. A method which recommends the best serve to a tennis player 
in a given context was developed by~\cite{wei2015}. 
Data mining technique were employed by~\cite{van2016} to discover patterns accounting for spatial 
and temporal aspects of volleyball matches to guide performance. 
\cite{brooks2016} proposed a soccer player ranking system based on the value of passes completed.

Our paper as well as all these other papers has the objective of understanding the dynamics of competitive 
sports to help the teams management, tactics development, providing more entertainment and profits.  

\section{Disentangling luck and skill}

In this section, our aim is to define a metric by which performance can be assessed with respect to a baseline where the teams have the same ability or skill and the tournament is determined by pure chance. It does not mean that all teams have the same 
probability of winning a given game as it depends on the context in which the match occurs. In soccer games, for example, 
the home team usually has a higher chance of winning. This is due to the presence of a favorable audience, and not 
because of any intrinsic skill the team may have. 
An analysis of the empirical frequency of home wins, away wins, and draws in nine successive seasons of the Brazilian S\'{e}rie A soccer league shows that the probability of an away win is 
about half the home win (0.25 versus 0.50, respectively), with little variation in 
time. Other contextual variables could influence the outcome of a game. 
For example, \cite{chen2016predicting} consider the 
different winning probabilities the same male tennis player has depending on which type of field he plays, such as 
indoor or outdoor court or the type of court surface (Grass, Clay, Hard, Carpet). In this paper, we consider only the home/away 
context by lack of additional information, but our model is easily extended if there is more contextual information available. 

\subsection{A coefficient to measure skill and luck}

Let $X_h$ be a random variable representing the points earned by a team when it plays in a match at home.
Similarly, let $X_a$ be associated with the points earned when the team plays away. 
The probability distribution of $X_h$ and $X_a$ is sports-specific, depending on the scoring system 
associated with each match. In basketball, for example, there is no ties and each game results in one 
point for either team. Hence $X_h$ and $X_a$ are simply binary Bernoulli random variables.
In the case of soccer the situation is different: 
a team earns either 3, 1, or 0 points depending on its number of goals 
being larger, equal, or smaller than its competitor, respectively. 
Let $P_h$, $P_t$ and $P_a$ be the probabilities that a home team wins, a tie, and a home team loses, 
respectively, with the restriction that they are non-negative numbers and $P_h + P_t + P_a = 1$.
Hence, the random variable $X_h$ has a multinomial distribution with possible values 3, 1, or 0 
with probabilities $P_h$, $P_t$, and $P_a$, respectively. As a consequence, its 
expected value $\mathbb{E}(X_h)$ is equal to $\mu_{X_h} = 3 P_{h} + P_{t}$
%$\mu_{X_h} = P_{h} \times 3 + P_{t}\times1 +P_{a}\times0$ 
and the variance is $\sigma^2_{X_h}= 9 P_{h} + P_{t} - \mu^2_{X_h}$. 
Similarly, $X_a$ has mean and variance given by 
$\mu_{X_a} = P_{t} + 3P_{a}$ and $\sigma^2_{X_a}= P_{t} + 9 P_{a} - \mu^2_{X_a}$. 
%$\mu_{X_a} = P_{h}\times0 + P_{t}\times1 +P_{a}\times3$ and $\sigma^2_{X_a}=P_{h}\times0^2 + P_{t}\times1^2 +P_{a}\times3^2 - \mu^2_{X_a}$. 

In a round-robin tournament composed by $k+1$ teams, each one of them plays $2k$ times, once at home and once 
at the opponent field. Let $X_{hi}$ and $X_{ai}$ be the score earned by a team when playing his $i$-th 
game at home and away, respectively. Then, $ Y_{2k} = \sum_i (X_{hi} + X_{ai})$.
is the total random score a given team obtains at the end of a tournament. Assuming that the probabilities 
$P_h$, $P_t$, and $P_a$ are the same for all teams is equivalent to assume that skills play no part in the tournament.
Any final score is the result of pure luck, mediated only by the contextual variables. The round-robin 
system, with one game at home and one away, and with the same number of games for all teams involved, 
guarantees that the distribution of its final score $Y_{2k}$ is the same for all teams. 
Considering the stochastic independence between the games and a sufficiently large number $k$ of games 
in a tournament, we have approximately a Gaussian distribution: 
$    Y_{2k} \sim N(\mu_{2k},\sigma^2_{2k}) $ 
where $\mu_{2k} = k(\mu_h+\mu_a)$ and $\sigma^2_{2k}=k(\sigma^2_h+\sigma^2_a)$. 

In summary, if all teams had the equal skill, after $k$ matches at home and $k$ away matches, the distribution 
of final scores $Y_{2k}$ follows approximately a normal distribution with parameters $\mu_{2k}$ and 
$\sigma^2_{2k}$. This represents our baseline model, against which the observed empirical results are to be contrasted. 
The construction of this distribution depends on the specific sport under analysis, 
on the number of games in the season and the amount of teams. 

In practice, the probabilities $P_h$, $P_t$, and $P_a$ are estimated from the data through the simple frequencies. 
For example, $P_h = W_h/N$ where $N$ is the total number of matches in a season and $W_h$ is the number of times 
there was a home win. Analogously, we estimate $P_t = W_t/N$ and $P_a = W_a/N$.

At the end of a season, we have the observed value of the final score $Y_{2k}$ for each team. 
This allows us to obtain a non-parametric, model-free estimate of its variance by calculating the empirical 
variance $s^2$. We compare the observed variability of $Y_{2k}$ relatively to its expected value under the equal skills
assumption: 
\begin{equation}
\phi = \frac{s^2-\sigma^2_{2k}}{s^2}
\label{eq:phi}
\end{equation}
The  value of $\phi$ lies between $[-\infty,1]$. Values around zero are compatible with seasons following the random model, 
where the teams have equal skills and luck is the only drive for victory. Positive values of $\phi$ indicate an excess 
variability in the final scores due to the additional variation induced by the different teams' skills. 
The closer to 1, the more dominant the skill factor. What can come as a surprise is the possibility of negative 
values for $\phi$. Values smaller than zero indicate that the teams' outcomes have a variability much smaller than 
that expected due to the luck factor. This could happen, for example, due to some kind of offsetting mechanism along the 
season or to some colluding scheme among the teams. Although this is not a common situation, we have clear evidence of 
these negative $\phi$ values presence in our data analysis (see Section \ref{sec:dataanalysisphi}).

\subsection{How much sampling variation in \texorpdfstring{$\phi$}?}

Since $\phi$ is an empirical measure based on the statistical outcomes in a season, 
even when the random model of no skills is true, it is virtually 
impossible to have $\phi$ exactly equal to zero, its expected value. %Pedro - Não entendi a frase abaixo:
We obviously need a measure of uncertainty to build confidence intervals in order to measure the distance between 
the observed data and the random model. 

We build a confidence interval for $\phi$ using Monte Carlo simulations under the random model. 
With the probabilities $P_h$, $P_t$, $P_a$, we generate Multinomial trials playing the teams against each other 
according to the same game schedule followed during the season. We calculate the simulated $Y_{2k}$ outcomes 
and their empirical variances. By repeating this independently, we obtain a 95\% Monte Carlo distribution 
for $\phi$ when there are no skill differences between the teams by taking the interval with endpoints given by 
the 2.5\% and 97.5\% percentiles. 

The confidence interval provide a yardstick to judge the value of $\phi$ calculated with the real data. 
If the observed  $\phi$ value is inside the interval, it is not significantly different of 0. This means that
there is no evidence that the league/season deviates from the random model. In contrast, a $\phi$ value above the 
upper interval endpoint is a strong indicator that the teams' skills are different, while $\phi$ values 
below the lower endpoint leads to the conclusion that the season has significantly less variability 
than the random model. 

Based on those seasons with $\phi$ values significantly different and higher than 0, a question of interest is: 
how many teams should be removed from the league/season in order to make it random? 
To answer this question, we ordered the teams according with their points in the season and remove the team more distant from the 
average team score. This team could be the best or the worst team in the season. In case of a tie between the best and the worst team, the best one is removed. If the tie is between two best teams or two worst teams, 
we use the alphabetical order to remove one of them. 
After selecting which team should be removed from the league/season, we recalculate the final score on the league/season excluding all matches between this team and the other teams. The $\phi$ value and the confidence interval are both recalculated. If the new $\phi$ value is inside the confidence interval, the algorithm stops. Otherwise, it keeps removing teams until the confidence interval contains the new $\phi$. 

\subsection{Where do these skills come from?}

%Pedro - ideia de contribuicao:
%a general framework to measure how skilled a team is. In order to do that, we first need to verify if skill plays a signicant role in the league. If it does, our framework identifies which teams are significantly more and less skilled than others. Finally, for leagues in which skill plays a significant role, we present a probabilistic graphical model to learn about the teams' skills.

As final part of this section, we present a probabilistic graphical model to learn about the teams' skills when 
their $\phi$ value is positive and outside the confidence interval. More important than simply learning the 
different skills levels of each team is to be able to factor them in terms of explanatory features. 
What are the most important factors to explain the different skills the teams show in a tournament? 
We propose a probabilistic graphical model that takes into account the features that lead to the 
variable skill as well as allowing for the presence of luck. 

A famous model for paired comparisons is the Bradley-Terry model~\cite{chen2009}. 
This model assumes that there are positive 
quantities $\alpha_1, \alpha_2, \ldots, \alpha_n$ representing the skills of $n$ teams. The probability 
that team $i$ beats team $j$ in a game is given the relative size $\alpha_i$ with respect to $\alpha_j$:
\[ \pi_{ij} = \mathbb{P}(i \mbox{ beats } j) = \frac{\alpha_i}{\alpha_i + \alpha_j} \: . 
\]
Take the positive parameters $\alpha_i$ to an unrestricted scale by transforming $\alpha_i = \exp(\beta_i)$. Then,
the Bradley-Terry formulation implies in a conveniently linear logistic model: 
\[ \log\left( \frac{p_{ij}}{1-p_{ij}} \right) = \beta_i - \beta_j  \: 
\]
The convenience of this linear logistic formulation is the possibility of extending the Bradley-Terry model to 
incorporate features that may help to estimate and to explain the differences between the $\alpha_i$'s skills. 

Let $\mathbf{x}_i$ be a $d$-dimensional vector with a set of features for team $i$ that may correlate with its 
skill $\alpha_i$. We want to learn the relevance of each feature present in $\mathbf{x}_i$ on the specification 
of the $\alpha_i$ skill. To do this, we need to specify a likelihood function for the random data holding the 
season outcomes. 

Most work using the Bradley-Terry approach adopt a binomial trial likelihood, considering only the final result, 
win or not, for a given game~\cite{chen2009}. 
We decided to take into account also the final score in the game as this should be 
correlated with the relative skills of the teams involved. 
Indeed, a large  $\alpha_i$ skill playing against a small $\alpha_j$ skill  
should not only have a large probability of winning but it should also lead to a greater score difference between the 
teams. We adapt the Bradley-Terry model for a Poisson likelihood model to account for the number of points or goals, 
rather than only the binary winning indicator.

\subsubsection{The likelihood}

In a tournament $K$ games and with $n$ teams having skills $\alpha_1, \alpha_2, \ldots, \alpha_n$, let $N_k$ be the total 
score of the home team and the away team at the $k$-th match during the season and $Y_k$ be the home team score in the same match. 
If we imagine each of the total $N_k$ points resulting in a success or a failure for the home team, the 
random variable $Y_k$ will have a binomial distribution conditioned on $N_k$ and on the teams skills. 
As the number of points can be very large (as in basketball games) and hence to produce a likelihood 
computationally intractable, the binomial distribution is approximated by a Poisson distribution with $N_k$
as an offset for the expected mean. 
We additionally allow unmeasured features and contextual variables to affect the distribution of $Y_k$ by 
introducing a random effect factor $\varepsilon_k$ for each match. 

In summary, conditioned on the parameters and random effects, each $k$-th game has 
\begin{equation}
  Y_k \sim \mbox{Poisson}\left( N_k ~ \frac{\alpha_{h(k)}}{\alpha_{h(k)} + \alpha_{a(k)}} + \varepsilon_k \right) 
\label{eq:distribY01}  
\end{equation}
where $h(k)$ and $a(k)$ are the indices of the home and away teams participating in the $k$-th season game. 

The presence of $\varepsilon_k$ is essential. Without it, the outcome variability induced by the Poisson distribution 
would be completely determined by the skill themselves through the $\alpha$ terms. We need to add an overdispersion 
component, the $\varepsilon_k$ random effect, to account for the large variation in the games outcomes. 

%\begin{eqnarray}
%Y_k & \sim & \mbox{Binomial}\left( N_k, \frac{\alpha_{h(k)}}{\alpha_{h(k)} + \alpha_{a(k)}} \right) \nonumber \\ 
%    & \approx & \mbox{Poisson}\left( N_k ~ \frac{\alpha_{h(k)}}{\alpha_{h(k)} + \alpha_{a(k)}} + \varepsilon_k \right) 
%\label{eq:distribY01}
%\end{eqnarray}

The skill coefficients $\alpha_1, \alpha_2, \ldots, \alpha_n$ are fully determined by intrinsic characteristics 
of each team. We use the canonical link function associated with the Poisson distribution to obtain a log-linear model
\[ \log(\alpha_i) = \mathbf{w}^T \mathbf{x}_i  
\]

Let $\mathcal{D} = \{ \mathbf{x}_i, Y_k, N_k, \mbox{ for } k=1,\ldots,K, \mbox{ and } i=1, \ldots, n \}$
be the observed dataset. The likelihood of the parameters $\mathbf{w} \in \mathbb{R}^d$ and $\varepsilon_1, \ldots, \varepsilon_K$
for all $k=1, \ldots, K$ games in the league based on $\mathcal{D}$ is given by 
\begin{align*}
& L(\mathbf{w}, \{ \varepsilon_k \} | \mathcal{D}) \propto  \prod_{k=1}^K \left(N_k \frac{\alpha_{h(k)}}{\alpha_{h(k)} + 
   \alpha_{a(k)}} + \varepsilon_k  \right)^{y_k} \\ 
&  ~~~ \times \exp\left( - N_k \frac{\alpha_{h(k)}}{\alpha_{h(k)} + \alpha_{a(k)}} + \varepsilon_k  \right)  \\
\end{align*}

\subsubsection{Hierarchical Priors}

\begin{figure}[htb]
\centering
\begin{tikzpicture}
    % nodes
    \node[obs] (y) {$Y_{k}$};%
     \node[latent,above=of y,xshift=1.5cm] (a) {$\bg{\alpha}$}; %
     \node[obs,right =of a, xshift=1cm] (x) {$\mathbf{x}_i$}; %
     \node[latent,above=of a] (w) {$\mathbf{w}$}; %
     \node[latent,above =of y, xshift=-1.5cm] (sigma) {$\varepsilon_k$}; %
    % plate
    \plate [inner sep=.3cm,xshift=.02cm,yshift=.2cm] {plate1} {(y)(a)(sigma)} {for($k$ in $1:K$)}; %
    \plate [inner sep=.3cm,xshift=.02cm,yshift=.2cm] {plate2} {(x)(a)} {for($i$ in $1:n$)}; %    
    % edges
     \edge {w,x} {a};
     \edge{sigma}{y};
     \edge[double]{a}{y}
\end{tikzpicture}
\caption{Probabilistic Graphical Model for the Bayesian skills model. 
The number of teams is $n$ and $K$ is the total number of matches in the season.}
\label{grafo}
\end{figure}
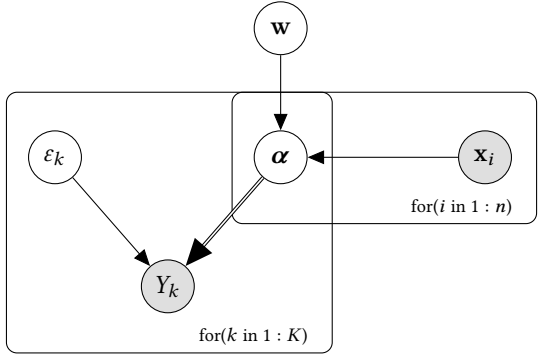

The prior distribution over the $\mathbf{w}$ weight components is a multivariate Gaussian distribution: 
$p(\mathbf{w}|\tau_w) \sim \mathcal{N}(\mathbf{w}; \mathbf{0}, \tau^{-1}_w \mathbf{I})$, where 
$\tau^{-1}_w$ is a precision parameter. We also assume independent prior 
zero-mean Gaussian distributions for the random effects $\varepsilon_1, \ldots, \varepsilon_K$.
Hence, the $n$-dimensional vector $\bg{\varepsilon} = (\varepsilon_1, \ldots, \varepsilon_n)$
also follows a multivariate Gaussian conditioned on a certain precision parameter: 
$p(\bg{\varepsilon} | \tau_{\varepsilon} ) \sim \mathcal{N}(\bg{\varepsilon}; \mathbf{0}, \tau^{-1}_{\varepsilon} \mathbf{I})$.
These prior distributions are the aspect of the Bayesian model that acts as a probabilistic regularization method. 
We completed the specification assigning Gamma hyper-prior distributions over the precision parameters as:
%\begin{eqnarray*}
%p(\tau_{w} | a, b) & \sim & \mathcal{G}(\tau_w ; a, b) \\
%p(\tau_{\varepsilon} | c, d) & \sim & \mathcal{G}(\tau_{\varepsilon} ; c, d) 
%\end{eqnarray*}
\[ p(\tau_{w} | a, b) \sim \mathcal{G}(\tau_w ; a, b) \quad \textrm{and} \quad p(\tau_{\varepsilon} | c, d) \sim  \mathcal{G}(\tau_{\varepsilon} ; c, d)  \]
where $\mathcal{G}(\tau,; a, b)$ is a Gamma distribution and $a$ and $b$ are global scale and rate hyper-parameters, respectively. Figure \ref{grafo} shows a representation of the Bayesian model developed in this paper.

\subsubsection{Bayesian inference via MCMC algorithms}

The Bayesian inference is based on the posterior distribution of the $\mathbf{w}$ weight coefficients, the random effects $\bg{\varepsilon}$, and $\tau_w$ and $\tau_{\varepsilon}$, the precision hyper-parameters. It 
is proportional to the product between the likelihood and the prior distribution:
\begin{align*}
& p(\mathbf{w}, \bg{\varepsilon}, \sigma_w^2, \varepsilon_k | \mathcal{D}) \propto  \prod_{k=1}^K \left( \frac{\alpha_{h(k)}}{\alpha_{h(k)} + 
   \alpha_{a(k)}} + \varepsilon_k  \right)^{y_k} \\ 
&  ~~~ \times \exp\left( - N_k \frac{\alpha_{h(k)}}{\alpha_{h(k)} + \alpha_{a(k)}} + \varepsilon_k  \right) 
        \exp\left( -\frac{1}{2 \sigma^{2d}_w} \mathbf{w}^T \mathbf{w} \right)  \\
& ~~~ \times \exp\left( -\frac{1}{2\sigma^{2K}_{\varepsilon}} \boldsymbol{\varepsilon}^T \boldsymbol{\varepsilon} \right) 
\end{align*}
where we eliminate all multiplicative factors not involving the unknown parameters.

The Metropolis-Hastings Algorithm, a member of Markov chain Monte Carlo class, 
is used to make inference about the coefficients in $\mathbf{w}$ 
by a sequential algorithm approach: 

\begin{lstlisting}
curr_coef = initial_values
for (k in 1:1000)
    for(n in 1:n_games)
        prop_coef = norm(curr_coef, sd = 0.15)
        if(prop_coef == accept)
            curr_coef = prop_coef
            
\end{lstlisting}

\section{Datasets} 

The 1503 seasons and their games studied in the first part of this paper were collected from the site \url{www.betexplorer.com}. The base has 270713 matches from 198 leagues occurred from January 2007 to July 2016. The games took place in 84 different countries from America, Europe, Asia, Africa and Oceania. 
We restrict the data collection to those sports leagues adopting a double round-robin  tournament system. This is a scheduling scheme where each team plays against every other team at least twice. A round-robin system provides a certain statistical stability for the results, preventing one single bad game from eliminating a team from the competition. It is also considered a fair system as it gives the same opportunities to all competitors. For our data analysis, it has the advantage of equalizing the number of matches in a tournament for all teams. The final ranking is determined by the accumulation of scores obtained along the competition. In our data collection, we also require
a tournament to have more than 7 teams and more than 5 seasons for statistical stability purposes. We collected data from basketball (42 leagues, 310 seasons), volleyball (51 leagues, 328 seasons), handball (25 leagues, 234 seasons) 
and soccer (80 leagues, 631 seasons).  The information about the NBA teams used in the second part the paper were collected from \url{www.basketball-reference.com}. From this site we collected the NBA players and their teams since 2004, the players salaries and the Player Efficiency Rating (PER) in each season.  

\section{Results}
\label{sec:results}

\subsection{The skill coefficient \texorpdfstring{$\phi$ }.} \label{sec:dataanalysisphi}

\begin{figure}[!h]
    \includegraphics[scale=0.40]{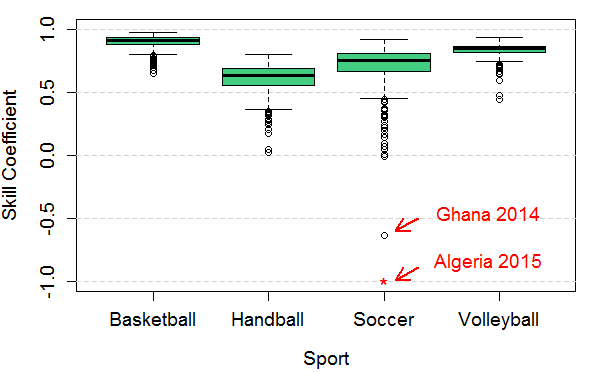}
    \caption{Boxplot of the 1503 seasons $\phi$ values separated by sport.}
    \label{phi_value}
\end{figure}

We calculated our skill coefficient $\phi$ for all seasons and all sports in our database. 
We show the results aggregated by sports in Figure \ref{phi_value}. 
The more compressed towards the upper limit 1, the more away from the random model and the more important is the skill component 
to explain the final scores. Basketball appears as the sport 
where the skill has the largest influence in the final results. 
In second comes volleyball, followed by soccer and then, handball. 
This is in accordance with previous analysis using other methods~\cite{ben2006parity, numbersgame}. 
%tirei o "most competitive" pois não se se eh a mesma coisa que queremos dizer

Considering the structure of each sport, it is possible to explain these results. The players usually score many points in basketball and volleyball games. Due to the long sequence of relevant events, it is more difficult for a less skilled team to win a game out of pure luck. However, in soccer and handball, there are much less relevant events leading to points. 
In soccer, for example, the average number of goals per match is 2.62. This makes easier for a less skilled team to win a match due to a single lucky event. Even if a large skill difference is present in soccer, 
it has less opportunity to be revealed as compared to basketball and volleyball. 
%Table \ref{tab2} shows 
We evaluated for each sport which factor, luck or skill, has more influence in its leagues results. If the observed value $\phi$ is not significantly different from zero (according to the confidence interval), we classified the corresponding season as random. Otherwise, it is classified as ``skill''. All basketball leagues and 99.39\% of volleyball seasons have a strong skill component in their final results. 
In contrast, in handball and soccer, pure luck can be the single factor in as many as 
17.95\% and 7.13\% of the seasons, respectively.  

\begin{comment}
% latex table generated in R 3.3.1 by xtable 1.8-2 package
% Wed Dec 14 17:48:22 2016
\begin{table}[!h]
\centering
\caption{Breaking down the calculated $\phi$ values by sports and luck or skill components.}\label{tab2} 
\begin{tabular}{l|cc|cc|c}
  \hline
  \multirow{2}{*}{Sport}&\multicolumn{2}{c|}{Skill}
&\multicolumn{2}{c|}{Random}& \multirow{2}{*}{Total}\\ \cline{2-5}
& Freq. & \%& Freq. & \% & \\ \hline\hline
Basketball & 310 & 100\% & 0 & 0\% & 310 \\ 
  Handball & 192 & 82.05\% & 42 & 17.95\% & 234 \\ 
  Soccer & 586 & 92.87\% & 45 & 7.13\% & 631 \\ 
  Volleyball & 326 & 99.39\% & 2 & 0.61\% & 328 \\ \hline
  Total & 1414 & 94.08\% & 89 & 5.92\% & 1503 \\ 
   \hline
\end{tabular}
\end{table}
\end{comment}

The values of $\phi$ for basketball and volleyball in Figure \ref{phi_value} are close to its maximum value 
and this could mislead one into thinking that luck plays no role in these sports. If this was so, these
sports would loose much of their appeal~\cite{Owen2013}. Skill has a strong influence in the results 
but it does not completely  determine the final ranking. In each season, some teams have more extreme 
skills while the others have approximately the same skill level. It is that first group that drives 
the $\phi$ coefficient towards and close its maximum value. Were they absent, and the season would resemble a lottery. 

\begin{figure}[!h]
    \centering
    \includegraphics[scale=0.35]{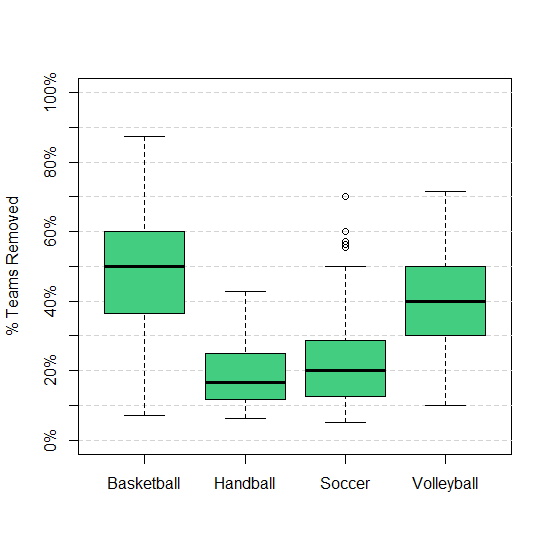}
    \caption{Percentage of Teams Removed.}
    \label{percent}
\end{figure}

To demonstrate that, in~Figure \ref{percent} we show the percentage of teams that should be removed from the league/season to make it random. The basketball (50\%) and volleyball (40\%) are the sports requiring relatively more teams to be removed. This reflects the results of Figure \ref{phi_value}.  Handball (14\%) and soccer(19\%), having typically lower values of $\phi$, need to remove relatively less teams 
to end up with a random league. To exemplify how different these sports are, consider two important soccer leagues, 
the Spanish \textit{Primera Divisi\'{o}n} and the English Premier League. In each season from 2007 to 2016, 
the average number of teams that need to removed was 3.2 and 4.9, respectively. As these leagues have 20 teams in the 
competition, these averages mean 16\% and 25\%, respectively. More important, the teams to be removed are almost the 
same every season. For example, in the Spanish league, the 3.2 average teams to be removed included Real Madrid 10 times 
and Barcelona 9 times out of the 10 seasons considered. This means that, removing 2 teams out of 20 in the  \textit{Primera Divisi\'{o}n}
we end up with a competition that produces a ranking that is essentially a random shuffle among the remaining teams. 
In the English Premier League, Manchester United appeared 9 times out of 10 seasons with other teams appearing as often as 7 times
in the list of teams that, if removed, renders the tournament basically a random outcome.  Consider now the NBA basketball competition. Reflecting the large skill differences among the teams, we need to remove between 17 to 25 teams out of 30 to 
make it look as a random competition. Also, the teams to be removed are highly variable from season to season, in contrast with soccer. 

\begin{figure*}
    \centering
    \includegraphics[scale=0.5]{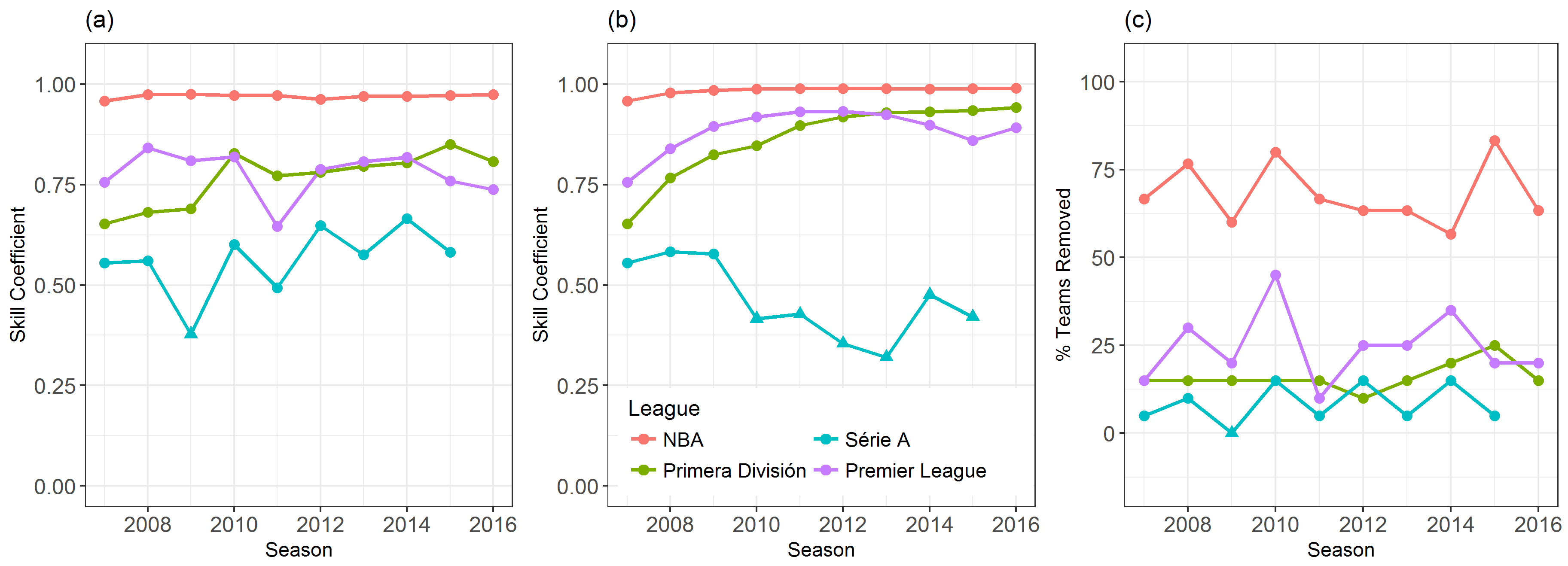}
    \caption{(a): Value of $\phi$ by season; (b) Cumulative $\phi$; (c) \% to be removed to make the league random.}
    \label{moreseason}
\end{figure*}

The panel (a) in Figure \ref{moreseason} shows the $\phi$ coefficient by season for some popular and financially important sport leagues. There is some slight trend but $\phi$ stays essentially  stationary around a sport-specific constant.
For NBA, $\phi$ is always above 0.95. The value for Premier League stays around 0.77,
while Primera Divis\'{o}n and S\'{e}rie A end up around 0.80 and 0.63, respectively. 

We consider these 4 leagues to study how the $\phi$ coefficient changes when the matches 
are accumulated over the seasons as if they comprise a single tournament.
At a given year $t \leq 2016$, we consider all matches occurred since 2007. 
Hence, the $\phi$ coefficient for, say, the 2009 season will consider now all matches between the teams that played in the 2007, 2008, and 2009 seasons
as if they were a single championship. 
The 4 leagues selected were the \textit{NBA} basketball league (from USA) and three soccer leagues: \textit{Premier League} (from England),
\textit{Primera División} (from Spain) and \textit{S\'{e}rie A} (from Brazil). The results are shown in panel (b) 
of Figure \ref{moreseason}. 
There was an increasing trend in the $\phi$ coefficient as the matches are accumulated, which implies that the relative skills of the teams are stable over the years, i.e., the point difference between good and bad teams is increasing. On the other hand, for the Brazilian soccer league the $\phi$ is decreasing, meaning that, in the long term, the teams are more similar to each other, with no small number of teams dominating the entire league, such as the Barcelona and Real Madrid 
do in the \textit{Primera Divisi\'{o}n} in almost all seasons.

The third panel (c) in Figure \ref{moreseason} shows the proportion of teams that need to be removed in each season 
for the league final ranking turn out into a completely random shuffle of the teams. The percentages show 
little evidence of trending during this period. NBA is a league where there is a large spread in skills,
requiring the removal of up to 62\% of the teams to make it random. The other three soccer leagues requires less,
with the English competition around 25\%, the Spanish being around  12\% and the Brazilian S\'{e}rie A requiring 
around 8\%. 

The most extreme $\phi$ coefficient in Figure \ref{phi_value} was observed in the soccer league Division 1 from Algeria, 
being equal to -1.93, in the season 2014-2015. Negative values are not very common and this was an extreme value, 
with $\phi \approx -0.6$ being the second smallest
value. The Algeria outlier value is only figuratively represented in the plot, in a vertical height different from its true vertical
coordinate, in order to a better visualization of the results. On this league and this year, the teams' scores were very similar 
and all teams still had chances to win the championship at a point very close to the season ending.
A newspaper headlined ``Algerian League is so tight all 16 teams can mathematically still win the title with 
four rounds of matches to go''~\cite{argelia}. 
During the third week, Albert Bodjongo, 
an athlete playing for the JSK team, was dramatically killed. 
At the end of a home game between JSK and USM Alger, when 
JSK lost by 2 to 1, angry JSK fans hit Bodjongo's head with a projectile thrown in the field. The athlete died 
a few hours later in a hospital. As a consequence, the Algerian Football Federation 
suspended all football indefinitely and ordered the closure of the stadium where the incident took place. 
The season was eventually resumed later. It is not clear how much this traumatic events can provide an explanation for the extremely negative value reached by $\phi$ in this case. 
%A much less competitive tournament may have ensued this unfortunate interruption. 

%\begin{figure}
%    \centering
%   \includegraphics[scale=0.4]{imagem3.png}
%    \caption{Newspaper headline showing how unusually similar were the 16 teams' scores 
%              in the 20142015 Algerian % soccer season.}
%    \label{argelia}
%\end{figure}

\subsection{Skills in the NBA}

We choose a league in particular, the National Basketball Association (NBA),  to fit our Bayesian 
model due to its feature data availability, prestige and high value of the $\phi$ coefficient in most seasons. 
The NBA is divided into two parts: the regular season and the playoffs. We considered only the matches during the 
regular season, because at this stage all teams played with each other at least twice (at home and away). 
As basketball does not have ties, the teams either earns one point if it wins, and zero points in case it loses. 
The seasons have 30 teams divided in East and West conferences. 

We have two types of features in the vector $\mathbf{x}_i$ to model the differences in skills. 
A subset of them are statistics associated with the teams, such as the average of the 5 highest players salaries. 
Other features are built based on a network that consider players and teams as in~\cite{vaz2012forecasting}. This network is represented by a graph where players and teams are nodes. The graph varies with year. In a certain year, two players are connected by an edge if they played together  sometime during the previous 6 seasons and if they both still play on NBA. A player and a team are connected by an edge if the player played on the team at some moment in the previous 6 seasons and if he still plays on NBA, either on that team or in another team. The complete 
list of features in $\mathbf{x}_i$ associated with team $i$ in a given season is the following:

\begin{adjustwidth*}{1em}{0em}
\noindent \textbf{CO}: A binary indicator. It is equal to $E$ if the team belongs to East conference, and $W$, if West conference; \newline
\noindent  \textbf{A5}: The average salary of the top 5 players salaries; \newline
\noindent \textbf{A6-10}: The average salary of the 6 to 10 top salaries. \newline
\noindent  \textbf{SD}: The standard deviation for the players salaries. \newline
\noindent  \textbf{AP}: The average Player Efficiency Rating (PER) considering the players in the last year. 
    The PER was created by John Hollinger and it is a system to rate the players according to their performance. The league-average PER is set 
    equal to 15.  \newline %cite http://www.basketball-reference.com/about/per.html
\noindent  \textbf{VL}: Team Volatility, measuring how much a team change its players between seasons. 
    It is defined as $\Delta d_t^y=d_t^y-d_t^{y-1}$, where $d_t^y$ is the degree of team $t$ in the year $y$.  \newline 
\noindent \textbf{RV}: Roster Aggregate Volatility, measuring how much the players have been transferred from other teams in the past years. 
    It is defined as $\sum\Delta d_t^y = \sum_{\forall v\in R^y_t } d_v^t / (y-y0_v)$, where $y0_v$ is the year of the first season of player $v$, where $v$ is a player of team $t$ in year $y$. $\sum\Delta d_t^y$ it the sum of the degree of all team members in the roster $R^y_t$ of the team $t$ in year $y$. \newline
\noindent  \textbf{CC}: Team Inexperience, the graph clustering coefficient. \newline
\noindent  \textbf{RC}: Roster Aggregate Coherence, measuring the strength of the relationship among the players. A high value of $\hat{cc}_t^y$ means that the team roster played together for a substantial amount of time or that few changes have occurred. The metric is defined as $\hat{cc}_t^y = avg(cc_v^t\times(y-y0_v)),\forall v \in R_t^y$, where $cc_v^t$ is the clustering coefficient of player $v$ in year $t$. \newline
\noindent  \textbf{SI}: Roster Size, the number of players. 
\end{adjustwidth*}

All variables are standardized to have mean zero and standard deviation 1. 
An extensive search was considered combining the features and the final model was selected 
using the \textit{Deviance Information Criterion} (DIC) \cite{spiegelhalter2002bayesian}, which 
balances goodness of fit and model complexity. When several models are compared, 
 that with the lowest value of DIC is the preferable one. 

We ran 10000 iterations of the Metropolis-Hastings and used a burn-in equal to 2000. 
The average acceptance rate of the Metropolis-Hasting Algorithm proposed new parameter values was 0.395,
a reasonable value, and the standard deviation was equal to 0.02. % ?? standard deviation do que ??
\begin{table*}[!h]
    \centering
        \caption{DIC value for some of the best models by season.}
    \begin{tabular}{cl|c|c|c|c|c}
    \hline 
    \multicolumn{2}{c|}{\multirow{2}{*}{Model Features}} 
    &  \multicolumn{5}{c}{DIC}\\ \cline{3-7}
        &  & 2012 & 2013 &2014 &2015 &2016\\ \hline
    1 &CO+A5+AP+VL+RC+SI & \textbf{-683589.7} &\textbf{-871853.9} &\textbf{-901784.2} &\textbf{-890471.0} &\textbf{-922389.1}\\
    2 &CO+A5+A6+AP+SD+VL+RC+SI & -678306.4& -865637.7& -895432.8& -882730.4& -915692.8\\
    3 &CO+A5+AP+SD+VL+RC+SI & -681779.5& -869800.7& -899301.4& -888251.6& -920525.5\\
    4 &CO+A5+A6+AP+VL+RV+CC+RC+SI & -675140.2 &-860787.3 &-890335.5 &-87362.2 &-908063.3\\ \hline
    \end{tabular}
    \label{dic}
\end{table*}

\subsection{Results}

The Table \ref{dic} presents the DIC value for some of the best models. 
It is possible to conclude that model 1, including the variables
$CO+A5+AP+VL+RC+SI$, is the best one in every single year we analysed. Variables $VL$ and $RC$ were also relevant 
in Vaz-de-Melo~\cite{vaz2012forecasting}. The variables $CO$, $A5$, and $AP$ were not present in~\cite{vaz2012forecasting}, 
but our analysis show that they have influence on the teams' skill in addition to that caused by $VL$ and $RC$.

\begin{figure*}[!h]
    \centering
    \includegraphics[scale=0.48]{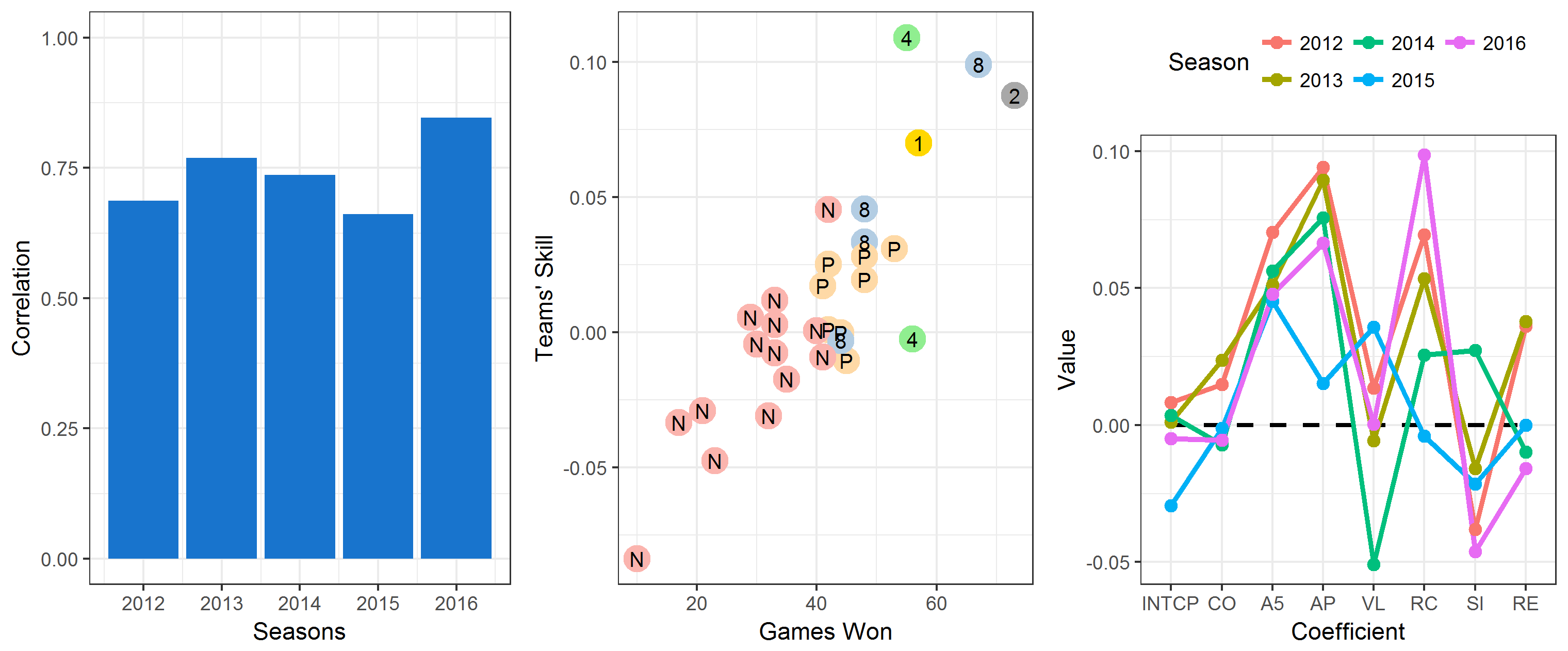}
    \caption{Barplot correlation between estimated skills and number of wins by season; 
    scatter plot of estimate skills versus number of wins for the NBA 2016 Season; 
    $w$ coefficients for different seasons.}
    \label{model}
\end{figure*}

Figure \ref{model} displays three plots with the results of model 1. The bar plot shows the correlation between the teams skills estimated by the Bayesian model (their posterior mean) and the number of games won during the regular season. The average over the seasons is equal to 0.7399, a high value. 
We can see from these plots that the high skills lead to a larger number of wins but this is not deterministic. This high correlation can be 
appreciated in the second plot, showing the scatter plot between the teams skills estimates $\hat{\alpha}_i$ and the games won by the $i$-th team in the 2016 season. Each circle is a team. The symbol inside the circle represent their position after the games in playoff: numbers are the ranking order, and the letters P and N represent the teams that lost on the first playoff game and the teams that did not play the playoffs, respectively. This scatter plot shows that our skills 
estimates $\hat{\alpha}$ based on the regular season data correlate very well with the total number of 
wins at the end of the playoffs. 
However, some surprising results can occur due to luck and additional events during the playoff season. 
Hence, the final ranking does not have  a perfect linear relationship with the number of wins. For example, the team with the highest skill estimated by the model ended on the $4$-th position in the final rank; the team with the second highest skill value was the second team in number of victories but it ended up only in the $8$-th position. The champion had the 4-th highest skill estimate and the 4-th highest number of games won. The third plot on Figure \ref{model} presents the coefficients estimated by our model 1 in each season. The average salary of 5 highest salaries (A5) and the average PER (AP) have a positive effect on skill, as one might expect.  Except for the 2015 season, the Roster Aggregate Coherence (RC) also has a positive effect. On the other hand, except for the 2014 season, the teams' size (SI) has a negative effect. 

%ou graficos 2, 3 e 4

Table \ref{tab:underwins} shows an important result. Based on the estimated skills $\alpha_i$, for each match, 
we are able to identify which team is the more skilled. Hence, looking at the frequencies classified according to 
the relative skills $\alpha_{h(k)}$ and $\alpha_{a(k)}$, we can estimate the probability $\mathbb{P}(U)$
that the underdog wins. It is stable over the seasons and approximately equal to $0.36$. This is the component 
of luck in the NBA outcomes. Our graphical model model allows us to be much more refined than this. 
We also estimate the probability that the underdog wins given that it plays away from home ($\mathbb{P}(U|A)$) 
and at home ($\mathbb{P}(U|H)$). We see that, being an average value, 
the unconditional probability is between these two conditional probabilities. Playing at home rather than 
away increases the underdog probability in approximately $0.45 - 0.27 = 0.18$, a substantial advantage.

% latex table generated in R 3.3.2 by xtable 1.8-2 package
% Wed Feb 15 23:14:53 2017
\begin{table}[!h]
\centering
\caption{Unconditional and conditional probabilities that an underdog team wins.}
\label{tab:underwins}
\begin{tabular}{c|ccc|cc}
  \hline
Season & $\mathbb{P}(U)$ & $\mathbb{P}(U|A)$ & $\mathbb{P}(U|H)$ 
     & $\mathbb{P}(U|A, R+)$ & $\mathbb{P}(U|H, R+)$  \\ 
  \hline
2012 & 0.35 & 0.27 & 0.44 & 0.25 & 0.24 \\
2013 & 0.36 & 0.25 & 0.47 & 0.16 & 0.13 \\
2014 & 0.37 & 0.29 & 0.45 & 0.15 & 0.14 \\
2015 & 0.37 & 0.30 & 0.44 & 0.22 & 0.20 \\
2016 & 0.34 & 0.26 & 0.43 & 0.19 & 0.13 \\ \hline 
Mean & 0.36 & 0.27 & 0.45 & 0.19 & 0.17 \\ \hline
\end{tabular}
\end{table}

We also consider what is this underdog winning probability when the adversary team is one of those 
removed by our $\phi$ coefficient to turn out the league random. We consider only those removed and among 
the best skilled teams. We denote them as $\mathbb{P}(U|A, R+)$ and $\mathbb{P}(U|H, R+)$ in Table 
\ref{tab:underwins}.  The previous probabilities should obviously decrease. Indeed, this is what 
the table shows. The home advantage almost disappears in this case. 
The proportions have more variability because they are calculated under a smaller 
number of games.

\section{Conclusions}

The proposal of this paper was to study the relative roles of skill and luck in some competitive sports. 
Through the $\phi$ coefficient, it was possible to determine that the basketball is the most competitive sport, in comparison with volleyball, soccer and handball. Despite volleyball final score be sets, that are limited to 5 in a match and maximum 3 by team, the long sequence of points during the match to obtain those sets made this sport more competitive than soccer and handball. It was also possible to find which teams should be removed from a league/season in order to make it random. In basketball and volleyball, in average one needs to remove 50\% and 40\%, respectively, while in soccer removing only 20\% turns the season in a random tournament.   

For leagues where the coefficient $\phi$ is positive and outside the confidence interval, we proposed a probabilistic graphical model to learn the teams' skills. We test this model in the NBA league from USA, due to the large amount of data available and to its $\phi$ coefficient being close to 1. Using the features conference, average salary top 5, average PER, team volatility, roster aggregate coherence and team size we found an average correlation of 0.7399 between the teams' skill estimated and the number of games won during the regular season.

Our graphical model allows us to decompose the relative weights of luck and skill in a competition. 
We found that, in the NBA seasons, 35\% an underdog team wins a match, representing the luck component 
in the NBA final scores. We also estimated that the home advantage of an underdog team adds 0.18 with respect to 
the average probability of 0.27 when it plays away from home. 

Looking at the most relevant features, our results suggest a management strategy to maximize the formation of teams with high skills, the controllable aspect on the team performance. Building smaller teams allow for higher salaries,
which also increases the odds of attracting high PER players and increasing rapport (roster aggregate coherence). 
Smaller, more cohesive teams have less conflicts and so are easier to manage.  

The final message of our paper is that sometimes the culprit is luck, about 35\% of the times in NBA. It is hard to beat luck in sports. This makes the joy and cheer of crowds. Or someone will forget the Superbowl 2017 evening?

\section{Acknowledgments}

The authors want to thank CNPq, CAPES, and FAPEMIG, for providing support for this 
work. 

\bibliographystyle{ACM-Reference-Format}
\bibliography{sigproc}

\end{document}